\pdfoutput=1

\documentclass[11pt]{article}

\usepackage[]{acl}

\usepackage{times}
\usepackage{latexsym}
\usepackage{graphicx}
\usepackage{tabularx}
\usepackage{microtype}
\usepackage{multirow}
\usepackage{latexsym}
\usepackage{booktabs}
\usepackage{courier}
\usepackage{amsmath}
\usepackage{subfig}
\usepackage{mdwlist}
\usepackage{placeins}
\usepackage{tikz}
\usepackage{hyperref}
\usepackage{lipsum}
\usepackage{etoolbox}
\usepackage{comment}

\usepackage[T1]{fontenc}

\usepackage[utf8]{inputenc}

\usepackage{microtype}

\newcommand{\todo}[1]{}
\newcommand{\jbgcomment}[1]{}

\title{Picard understanding Darmok: A Dataset and Model for Metaphor-Rich Translation in a Constructed Language}

\author{Peter A. Jansen \\
  University of Arizona \\
  \texttt{pajansen@arizona.edu} \\\And
  Jordan Boyd-Graber \\
  University of Maryland \\
  \texttt{ jbg@umiacs.umd.edu} \\}

\begin{document}
\maketitle
\begin{abstract}
Tamarian, a fictional language introduced in the \textit{Star Trek} episode \textit{Darmok}, communicates meaning through utterances of metaphorical references, such as \textit{``Darmok and Jalad at Tanagra''} instead of \textit{``We should work together.''} This work assembles a Tamarian-English dictionary of utterances from the original episode and several follow-on novels, and uses this to construct a parallel corpus of 456 English-Tamarian utterances.  A machine translation system based on a large language model (T5) is trained using this parallel corpus, and is shown to produce an accuracy of 76\% when translating from English to Tamarian on known utterances.\footnote{Data and code available at: \url{https://github.com/cognitiveailab/darmok}}
\end{abstract}

\section{Introduction}

\noindent Science fiction and fantasy literature has long created constructed languages for their characters, from Elvish in \textit{Lord of the Rings} and Klingon in \textit{Star Trek} to Heptapod in \textit{Arrival}~\cite{cheyne-08}.  These languages often have many of the same syntactic or semantic features as human languages, and some (such as Klingon) have been developed to a level where full dictionaries \cite{okrand1992klingon} and online translators are available.\footnote{https://www.translate.com/klingon-english}

An unconventional language was proposed in an episode of \textit{Star Trek: The Next Generation} called \textit{``Darmok''}, where a race of aliens called the Tamarians speak a language that is communicated exclusively through metaphors.  Instead of direct reference (e.g. \textit{``I want to give this to you''}), Tamarians speak in metaphorical references grounded in stories (e.g. \textit{``Temba, his arms wide''}) that (like symbols) have learned associations with their true meaning meaning.  In the \textit{Darmok} story, the unusual nature of the language poses a challenge for both the automated translation systems and the characters in the story to learn.
The creator of the language, Joe Mendowsky was inspired by the difficulty of translating across cultures~\cite{block-12}, and Tamarian has since been the subject of repeated informal study \cite{bogost_2014} in the 30 years since the episode aired. 

\begin{figure}[!t]
	\centering
	\includegraphics[scale=0.85]{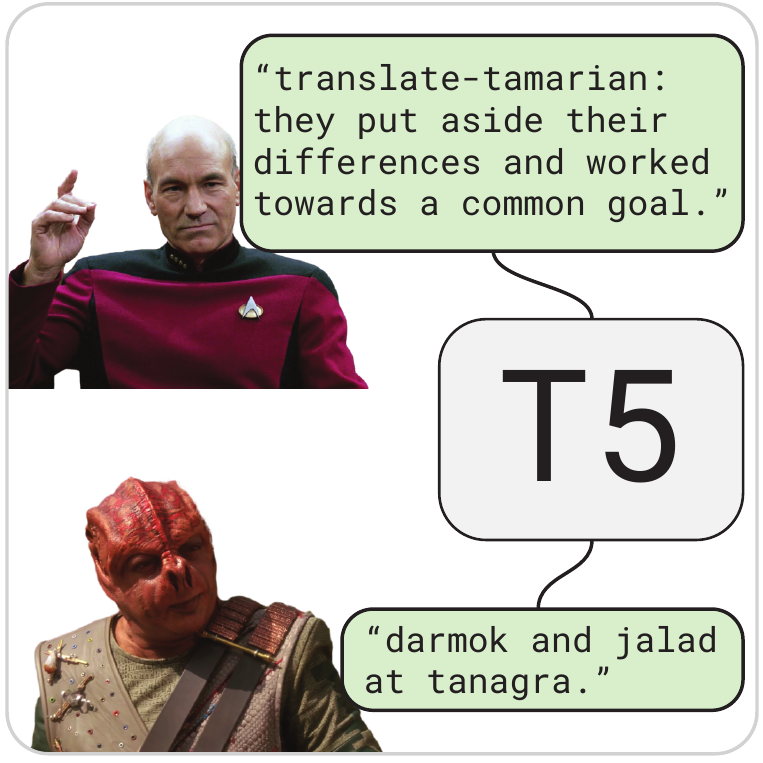}
	\caption{An example of translating English to the metaphor-grounded Tamarian language using T5.}
	\label{fig:example}
	\vspace{-4mm}
\end{figure}

This work investigates the feasibility of translating this artificial metaphor-rich language via our new parallel corpus of English-Tamarian phrases (Figure~\ref{fig:example}).  Our machine translation system based on a large language model \cite[T5]{raffel2020exploring} has 76\% accuracy in translating English phrases to Tamarian metaphorical utterances.  This suggests automatically translating metaphor-grounded languages may be feasible, though we discuss several pragmatic challenges in representing complex expressions and generating a parallel corpus preventing scaling the approach. 

%
%

%
%
\begin{table*}[t!]
\begin{center}
\footnotesize
\begin{tabular}{cllp{6.5cm}} 
\toprule
 ~ & Tamarian Utterance  & Inferred Meaning &   English Example \\
 \midrule
 1  &   Darmok and Jalad at Tanagra     & Working together  &   Knowing they would both be needed, they went together. \\
 2  &   Temba, his arms wide.           & Giving            &   The child offered his toy to his friend. \\
 3  &   Kira at Bashi.                  & Story-telling     &   They described what had happened to those who listened.\\ 4  &   Chenza at court, the court of silence.  &   Incontestability    &   The results were beyond reproach. \\  
 5  &   Zima at Anzo, Zima and Bakor.   &   Persistence     & They continued their task, undeterred from past failures.\\
 6  &   Fendit, refusing the flame.     & Refusing help     & She preferred to work alone, without assistance.\\
 7  &   Chatha and Teribium, the fire warm. & Hospitality   & Their household was offered for rest and comfort. \\
 8  &   Jeral, her arms weary.          & Being tired       & She was spent at the end of the day. \\
 9  &   Pirakee, with clouds parted.    & Visibility        &   She turned on a flashlight, making it easier to see. \\
 10 &   Hammat dancing.                 & Liking something  &   It filled them with delight.\\
\bottomrule
\end{tabular}
\caption{\footnotesize Example Tamarian utterances, their inferred meaning, and an English example from the parallel corpus. 
\vspace{-4mm}
\label{tab:corpus-examples}}
\end{center}
\end{table*}

\section{English-Tamarian Parallel Corpus}

Comparatively few Tamarian utterances have been authored, effectively limiting the size and scope of the effort.  To maximize the number of available utterances, all utterances from the original broadcast episode, as well as those in three licensed novels featuring a Tamarian main character were used \cite{beyer2012,beyer2014,beyer2015}. Approximately twenty utterances are provided in the \textit{Darmok} episode, while an additional forty-eight are used in the novels, for a total of sixty-eight utterances. 

\paragraph{Tamarian-to-English dictionary:} To create a parallel English-Tamarian corpus, first a Tamarian-to-English dictionary that captures the inferred meaning of each Tamarian utterance was required.  The meanings of the twenty broadcast utterances was ascertained from a Reddit thread with extensive discussion of the topic.\footnote{\url{https://www.reddit.com/r/DaystromInstitute/comments/4ggwo5/the_tamarian_language_an_analysis/}} The meanings of the remaining forty-eight utterances was inferred as best as possible from the surrounding context of where they appeared in their respective novels.

\paragraph{Tamarian-English Parallel Corpus:} Training a machine translation system requires a parallel corpus, where utterances of one language are paired with utterances of a second language, where the utterances in both languages have the same meaning.  Tamarian utterances abstractly refer to specific types of situations that could be applicable to many circumstances.  Thus, for each Tamarian utterance a set of~$k$ English examples were manually authored, with ten examples authored for thirty-nine utterances, and five examples authored for eleven utterances.  Eighteen Tamarian utterances were not included in the parallel corpus as they have relatively narrow meanings, and generating a large number of parallel examples for them in English proved challenging.  The final parallel corpus contains fifty Tamarian utterances, paired with 456 parallel English utterances (Table~\ref{tab:corpus-examples}).

\section{Translation Model}

\paragraph{Approach:} Here, English-to-Tamarian is modeled as a sequence-to-sequence (seq2seq) learning task, using English utterances as the source sentence, and a single Tamarian translation of that English utterance as the target sentence.  


\paragraph{Models:} Modeling used T5 \cite{raffel2020exploring}, a large pre-trained multi-task language model.  T5 includes pre-training for a variety of tasks, including question answering, summarization, and translation.  Several model sizes were explored, including T5-small (66M parameters), T5-base (220M parameters) and T5-large (220M parameters).  The model prompt took the form of: 

\begin{center}
{\footnotesize\texttt{translate English to Tamarian: \textit{\{src\}}}}
\end{center}

\noindent where \textit{\{src\}} is the English source sentence to translate (e.g. \textit{``She offered it to them''}).  The model then generated a corresponding target sequence corresponding to the Tamarian translation of the source sentence (e.g. \textit{``Temba. His arms wide.''}). The model was implemented using the Huggingface Transformers library \cite{wolf2020transformers}.

\paragraph{Dataset splits:} Due to small dataset, we use 5-fold crossvalidation: with 60\% of data used for training, 20\% for development, and 20\% for test.  For utterances with ten examples, this corresponds to six train, two development, and two test samples per run, while for utterances with five examples, this corresponds to three train, one development, and one test sample per run. 


\paragraph{Evaluation Metrics:} Translation performance was evaluated using \textsc{sacreBLEU}~\cite{post-2018-call}, a metric that measures translation performance using $n$-grams, while taking partial matches into account. 
Here, because only fifty Tamarian utterances are available, and their surface presentation is generally constant, we also consider evaluating translation as an \textit{$N$-class} classification task where a given English input sentence can be classified as one of fifty Tamarian utterances. 

%
%
\begin{table}[t!]
\begin{center}
\footnotesize
\begin{tabular}{cccccc} 
\toprule
 ~ & ~       & \multicolumn{4}{c}{Translation Performance} \\
 ~ & ~       & \multicolumn{2}{c}{Dev.} & \multicolumn{2}{c}{Test} \\\cmidrule(lr){3-4}\cmidrule(lr){5-6}
 ~ & Model       &   BLEU   &   Acc.  & BLEU   &   Acc. \\
 \midrule
 \parbox[t]{2mm}{\multirow{2}{*}{\rotatebox[origin=c]{90}{T5}}} & 
     T5-Small   & 38    & 34.4\%    &   41  &   38.0\%      \\ 
 ~ & T5-Base    & 71    & 72.8\%    &   70  &   72.4\%      \\
 ~ & T5-Large   & \textbf{80}    & \textbf{82.4\%}    &   \textbf{74}  &   \textbf{76.4\%}      \\
\bottomrule
\end{tabular}
\caption{\footnotesize Average English-to-Tamarian translation performance on both development and test sets.  BLEU measures per-token accuracy, while \textit{Acc.} refers to the average binary classification accuracy  of choosing the correct Tamarian utterance for a given English input sentence.}
\vspace{-6mm}
\label{tab:translation-performance}
\end{center}
\end{table}

\section{Results}
Models were trained until performance (BLEU) asymptoted on the development set, at thirty epochs.
The best performing model achieves a translation accuracy of 76\% on the unseen test set, which corresponds to translating approximately three out of four English utterances from the corpus correctly into Tamarian (Table~\ref{tab:translation-performance}).


\section{Discussion}

As a constructed language for a fictional universe, Tamarian is a low resource language with fewer than one hundred known utterances.  What might it take to grow Tamarian (or a metaphorically-grounded Tamarian-like language) into a more complete artificial language similar to Klingon? 
This section attempts to address the challenges of scaling beyond this work in the context of two central difficulties: growing the parallel corpus of metaphors, and challenges associated with the semantics of translating complex ideas in Tamarian.  

\subsection{Growing the Parallel Corpus}

Growing the vocabulary of metaphors in Tamarian presents a unique challenge for constructed languages.  Where human languages typically expresses base-level semantics at the level of the morpheme or word, Tamarian's most atomic construction is a single metaphor, making approaches that start with translating a dictionary challenging to adapt.  One approach to growing Tamarian would be to continue the current manual approach, identifying a set of atomic events that convey common situations (such as \textit{eating, giving, taking, or helping}), and authoring utterances grounded in an expanded Tamarian mythology---for example, \textit{``Timba, his stomach rumbling''} to convey the notion of hunger.  The prerequisite for having an exhaustive list of possible event schemas to translate would likely make this approach challenging to scale.


\paragraph{Automatic Generation:} An alternate approach was suggested by Picard in \textit{Darmok} -- to use the existing body of human literature  (such as the \textit{Epic of Gilgamesh}) to build a Tamarian-like language grounded in metaphors inferred from classic literature. Picard suggests that \textit{``Gilgamesh and Enkidu at Uruk''} might be an utterance to represent a central component of the story -- two people who were first in conflict coming together in friendship.  Such an automatic approach to building a Tamarian-like language is in principle feasible, potentially making use of recent successes in automatic summarization to extract key elements of a story in templated form (e.g. \textsc{\{PersonX\} and \{PersonY\} at \{Location\}}) to generate novel utterances.  One of the challenges with this approach is that narratives often contain many events, specified both at a low-level (e.g. Enkidu entering the city of Uruk) and high-level (e.g. Gilgamesh and Enkidu eventually forming a friendship in spite of their differences), and identifying only a single idea to be represented by the utterance would be difficult. 

%
%
\begin{table}[t!]
\begin{center}
\footnotesize
\begin{tabular}{ll} 
\toprule
 \textbf{Tamarian Utterance}  & \textbf{Inferred Meaning} \\
 \midrule
 \multicolumn{2}{l}{\textit{Gesture/Context Hypothesis}}\\
 \midrule
 ~~~Temba, his arms wide.           & Hand me the blue screw- \\
 ~~~\textit{Concurrently: Pointing at item} & driver I am pointing at \\
 ~\\
 \midrule
 \multicolumn{2}{l}{\textit{Specificity Hypothesis}}\\
 \midrule
 ~~~Jeral, her gift.               & Give me a blue \\
                                   & screwdriver on the left  \\
~\\                                   
 \midrule
 \multicolumn{2}{l}{\textit{Modifier Hypothesis}}\\
 \midrule
~~~Temba, his arms wide.           & Giving           \\
~~~Paris, in the garage.           & Screwdriver      \\
~~~Tolanis painting, in winter.    & Blue             \\
~~~Bakor, examining.               & Look to the left   \\
\end{tabular}
\caption{\footnotesize Examples of the three hypotheses for how fine-grained semantics could be inferred or composed in Tamarian.
\vspace{-8mm}
\label{tab:hypotheses}}
\end{center}
\end{table}

\subsection{The Challenge of Translating Fine-grained Semantics}

It has been hypothesized that Tamarian may not be well suited to expressing fine-grained semantics, and would present challenges for translating utterances such as \textit{``Hand me the blue screw driver on the left``} \cite{bogost_2014}.  While the few observed multi-utterance exchanges of Tamarian have (so far) typically conveyed steps in a story, we present three hypotheses for how fine-grained semantics might be achieved, with examples shown in Table~\ref{tab:hypotheses}:
\begin{enumerate*}
    \item \textit{Gesture/Context hypothesis:} The spoken Tamarian language may ground ambiguity through gestures or other situated contextual cues, as the Tamarian captain does when he utters \textit{``Temba, his arms wide'' (take)} and gestures to a weapon. 
    \item \textit{Specificity hypothesis:} Though impractical, the Tamarian language may have many utterances to refer to very specific situations. 
    \item \textit{Modifier hypothesis:} Unobserved classes of utterances may serve as modifiers, providing additional clarification to an utterance. 
\end{enumerate*}
There is partial observation of both the \textit{gesture/context} and \textit{modifier} hypotheses provided in the original \textit{Darmok} episode, and we believe the modifier hypothesis likely provides a mechanism for composing larger units of meaning akin to a generative grammar.

\jbgcomment{Feel free to cut if you think this limitation is too negative}

The more fundamental challenge of extending Tamarian is that every sentence must be connected to an underlying mythology: if you want to translate a sentence you must first create a universe~\cite{sagan-83}.
While we can invent Tamarian sounding proper nouns, a more fundamental challenge is to build a world where there are characters who would have or invent a screwdriver, a character who could successfully use it, a character who would use it incorrectly, and perhaps someone else who could address when you've accidentally stripped the head of the screwdriver.

Thus, the challenge is not just creating enough examples but also building the cultural cannon to support those examples.
While this is a unique linguistic challenge for Tamarian, it follows the course of other constructed languages: Quenya was developed alongside the backstory of Middle Earth~\cite{lewis-95} and the creator of the Klingon language also ensured that the Klingon mythology was recorded in the Klingon language~\cite{Solr-doab03841}.
Tamarian foregrounds this challenge of obtaining enough cultural context to translate~\cite{keesing-85,maitland-17}.

\section{Related Work: (Computational) Linguistics for Constructed Languages}

The elephant in the room is whether it is worthwhile to study constructed languages at all.
This section seeks to answer that question with a resounding yes by discussing the other insights that have come from scholarly investigations of constructed languages.

Tamarian is from the Star Trek Universe, so it is instructive to spend a little time first with the oldest Star Trek language, Klingon.
Klingon is often used in \textsc{nlp} education because it has features that are rare in natural languages but it is incredibly regular: a morphological analyzer can get 100\% accuracy but still have fascinating properties like affixes for honorifics, completion, and tense~\cite{wicentowski-04}.
Likewise, because Klingon is by construction meant to feel literally alien, its \textsc{ovs} structure can also upend students' part of speech tagging expectations~\cite{boyd-graber-14:klingon}.

But Klingon is not just a fun exercise for programmers and linguists; the creation of parallel data (as discussed above for Tamarian) also explores the interplay between culture and translation.
For the translation of \textit{Hamlet} into Klingon, cultural adaptation~\cite{Peskov-21} is also needed: for example, Fortinbras becomes ``the most insuborinate head of the House of Duras''~\cite{Kazimierczak-10}.
The art of translation often relies on metaphor~\cite{veale-16} and cultural knowledge~\cite{vinay-95}, and just as exploring Klingon can reveal limitations of our understanding of affix morphology and \textsc{ovs} word order, Tamarian can help illuminate the limitations of metaphor in communication.

All extant constructed languages are low resource languages, which typically pose challenges for machine translation~\cite{haddow-21}.
Like how Klingon can emphasize particular aspects of a language (word order, morphology), Tamarian helps focus attention on the role of mythology, inter-personal relationships, and multiword expressions for translation.

\section{Conclusion}

This paper is an initial English--Tamarian translation model.  
This task is difficult because it not only maps words to words but also maps metaphor to typical translation phrases.  
While Tamarian is a constructed language, it shows large language models' ability and limitations for metaphor.



\bibliography{bib/journal-full,bib/pj,bib/jbg,bib/languageresource}
\bibliographystyle{acl_natbib}


\end{document}